\documentclass[journal]{IEEEtran}

\ifCLASSINFOpdf
\else
   \usepackage[dvips]{graphicx}
\fi
\usepackage{url}

\usepackage{graphicx}

\usepackage[capitalise]{cleveref}
\usepackage{subfigure}

\usepackage{multirow}
\usepackage{cite}
\usepackage{soul}

\begin{document}

\title{Progressive Class-based Expansion Learning For Image Classification}

\author{Hui~Wang,
	Hanbin~Zhao,
	and~Xi~Li
	\IEEEcompsocitemizethanks{\IEEEcompsocthanksitem H.~Wang, H.~Zhao, and X.~Li are with College of Computer Science and Technology, Zhejiang University, Hangzhou 310027, China. \protect
		
		E-mail: \{wanghui\_17, zhaohanbin, xilizju\}@zju.edu.cn. \protect
	}
	\thanks{(Corresponding author: Xi Li.)}
}

\markboth{Journal of IEEE Signal Processing Letters}
{Shell \MakeLowercase{\textit{et al.}}: Bare Demo of IEEEtran.cls for IEEE Journals}
\maketitle

\begin{abstract}
In this paper, we propose a novel image process scheme called class-based expansion learning for image classification, which aims at improving the supervision-stimulation frequency for the samples of the confusing classes. Class-based expansion learning takes a bottom-up growing strategy in a class-based expansion optimization fashion, which pays more attention to the quality of learning the fine-grained classification boundaries for the preferentially selected classes. Besides, we develop a class confusion criterion to select the confusing class preferentially for training. In this way, the classification boundaries of the confusing classes are frequently stimulated, resulting in a fine-grained form. Experimental results demonstrate the effectiveness of the proposed scheme on several benchmarks.
\end{abstract}


\IEEEpeerreviewmaketitle

\section{Introduction}

\IEEEPARstart{C}{onvolutional} neural networks~\cite{krizhevsky2012imagenet,simonyan2014very,szegedy2015going,he2016deep,huang2017densely} (CNN) have attracted considerable attention in image classification due to their effectiveness in representation learning~\cite{bengio2013representation,zhang2018network}. Since they require computationally expensive and memory-consuming operations, CNN training typically resorts to stochastic gradient descent (SGD)~\cite{robbins1951stochastic, rumelhart1988learning} for iterative batch-level learning, it traverses the entire training dataset across randomly generated batches throughout successive epochs. With this epoch-by-epoch learning procedure, the classification boundaries of the CNN model are dynamically updated until convergence. Due to the memory resource limit, the samples within a smaller-size batch usually distribute extremely diversely and sparsely, resulting in a low supervision-stimulation frequency for each sample. The low frequency for each sample in turn causes the learning process to pay more attention to the learning quality of the coarse-grained classification boundaries while ignoring fine-grained details. Therefore, seeking for an effective and stable image classification strategy remains a key issue to solve in the CNN learning area. 

\begin{figure}[t]
	\centering
	\includegraphics[width=1\columnwidth]{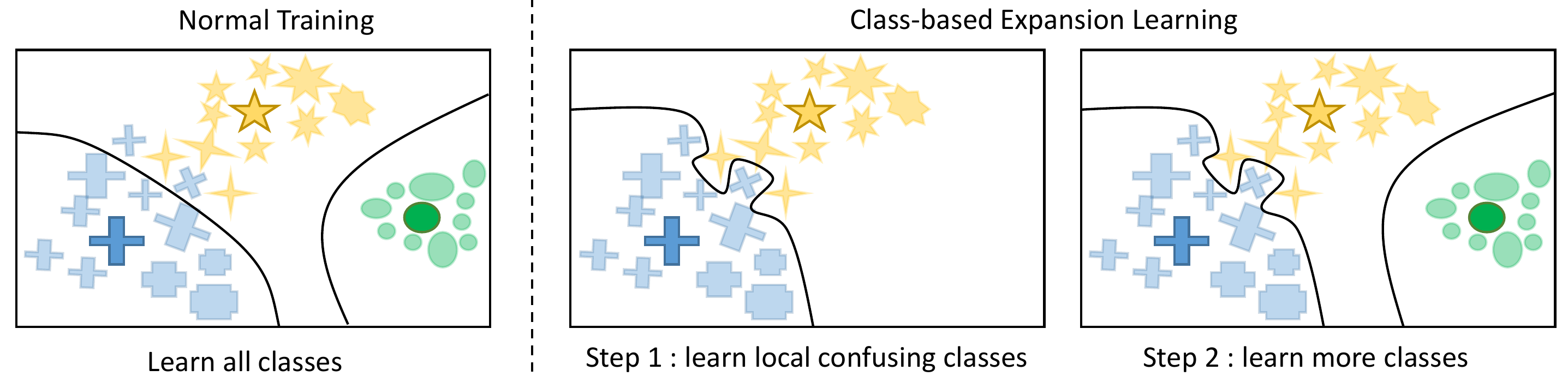}
	\caption{Examples of class-based expansion learning and normal training (Black curves in each figure denote the classification boundaries). Normal training pays more attention to the learning quality of the coarse-grained classification boundaries while ignoring fine-grained details. Our Class-based expansion learning pays more attention to fine-grained details of the classification boundaries for confusing classes.}
	\label{fig:cel}
\end{figure}

To date, curriculum learning~\cite{bengio2009curriculum,spitkovsky2010baby,basu2013teaching,graves2017automated,zhu2020curriculum,wang2021survey} and self-paced learning~\cite{kumar2010self,jiang2014self,jiang2015self,meng2017theoretical,gu2016robust,yu2018self,soviany2021curriculum} have been proposed to improve the speed of convergence of the training process to a minimum and the quality of the local minima obtained. The key concept of the proposed methods is inspired by human behavior, who always learn new things from ``easy'' to ``complex''. But these methods still do not consider the fine-grained classification boundaries.

In this letter, we propose a new learning pattern that arises from the inspirations of the biological learning mechanism. The Hebbian theory~\cite{hebb2005organization} delivers an important insight that the increase in synaptic effects of synaptic cells comes from repeated and sustained stimulation. Meanwhile, the human learning pattern usually follows a progressive knowledge expansion learning pipeline, it dynamically learns new knowledge while keeping the old knowledge frequently reviewed. The knowledge that is frequently reviewed is often better learned. Inspired by the biological learning mechanism, we propose a progressive learning pipeline that aims at effectively enhancing the supervision-stimulation frequency for each sample to enhance the quality of the fine-grained classification boundaries as shown in \cref{fig:cel}. Specifically, we present a progressive piecewise class-based expansion learning scheme, which first learns fine-grained classification boundaries for a small portion of classes and subsequently expands the classification boundaries with new classes added. Therefore, the presented class-based expansion learning scheme takes a bottom-up growing strategy in a class-based expansion optimization fashion, which puts more emphasis on the quality of learning the fine-grained classification boundaries for dynamically growing local classes. Besides, we propose a class confusion criterion to sort the classes involved in the class-based expansion learning process. The classes where the samples have large intra-class and small inter-class distances on average (i.e., class confusing samples) are preferentially involved in the class-based expansion learning process. Once a particular class is selected, all the samples belonging to this class are added to the training sample pool for the CNN model learning. Such an expansion procedure is repeated until all the class samples participated in the training process. In this way, the classification network model is dynamically refined based on the updated training sample pool, and the classification boundaries of the preferentially selected classes are frequently stimulated, resulting in a fine-grained form.

The main contributions of this work are summarized as follows: i) Motivated by Hebbian theory, we propose to investigate the influence of ``stimulation frequency'' on neural network learning and make an observation that the poor performance of the confusing classes is partially a result of the low stimulation frequency. ii) We propose a novel class-based expansion learning pipeline to deal with the learning problem. This pipeline progressively trains the CNN model in a hard-to-easy class-based growing manner, thereby the classification boundaries of the preferentially selected confusing classes are frequently stimulated.  iii) We develop two class confusion criteria to sort the classes for the class-based expansion learning process. Extensive experiments demonstrate the effectiveness of our work against conventional learning pipelines on several benchmarks.

\begin{figure}[t]
	\centering
	\includegraphics[width=1\columnwidth]{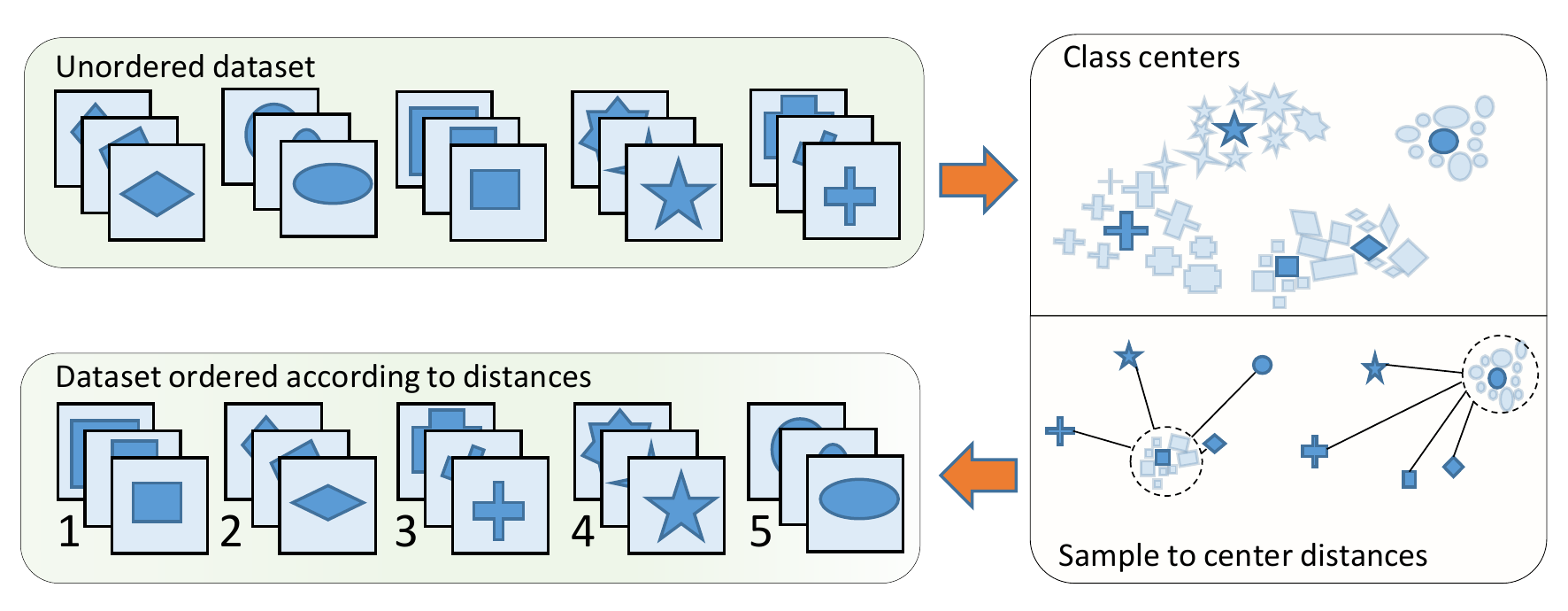}
	\caption{Illustration of our class confusion criterion. We calculate the confusion score of each class on an unordered dataset. The confusion score of a class is high when the samples of the class are far away from the corresponding same-class center and meanwhile close to other class centers (e.g. rectangle in this figure). Finally, we sort the classes of the unordered dataset in descending order of the confusion score and obtain an ordered dataset.}
	\label{fig:order}
\end{figure}
\section{Method}
In this section, we detail the proposed class-based expansion learning scheme. We first formally define the problem in \cref{problem_df}, we describe our algorithm to solve it in \cref{criterion,expansion}.
\subsection{Problem Definition}\label{problem_df}
%
Given a $M$-classes dataset $D=\{C_1,C_2,\dots,C_M\}$, and the $m$-th class of the dataset $C_m$ contains $N_m$ samples $x^m$ and their corresponding labels $y^m$:
\begin{equation}\label{dataset_class}
\begin{array}{cl}
C_m=\{(x_1^m, y^m), (x_2^m,y^m),\dots,(x_{N_m}^m, y^m)\}.\\
\end{array}
\end{equation}

Let $f(\cdot;\theta)$ denote the mapping function of the CNN model, where $\theta$ represents the model parameters inside $f(\cdot)$. In a typical training process, the goal is to learn an optimal $\theta^*$:
\begin{equation}\label{func}
\begin{array}{ll}
\theta^* = \mathop{\arg\min}\limits_{\theta} \sum\limits_{(x, y) \in D} l(f(x;\theta),y),
\end{array}
\end{equation}
where $l(\cdot,\cdot)$ is the loss function (e.g. cross-entropy loss).

\subsection{Class Confusion Criterion}\label{criterion}
In this section, we introduce our metric for deciding the order in which classes are presented to the class-based expansion framework. Ideally, we want to set the score of a class to a high value when it is easy to confuse.

We use a pre-trained tiny network $g(\cdot)$ to evaluate the confusion score for each class. Note the training cost of $g(\cdot)$ is much lower than that of $f(\cdot)$. To obtain the score of each class, we start by using $g(\cdot)$ to transform samples from image space into feature space and logits space:
\begin{equation}\label{fea}
\begin{array}{ll}
g^m_x = g_f(x^m)\\ 
p^m_x = g_c(g^m_x)
\end{array}
\end{equation}
where $g_f(\cdot)$ and $g_c(\cdot)$ denote the feature extractor and the classfier of the network $g(\cdot)$. 
Afterwards, we propose two kinds of class confusion criteria:

\noindent\textbf{Distance-based Criterion.} To obtain it, we first calculate the class center of each class:
\begin{equation}\label{class_center}
\begin{array}{ll}
u^{m}= \frac{1}{N_m}\sum\limits_{(x,y)\in C_m} g^m_x,
\end{array}
\end{equation}
where $N_m$ is the number of samples in class $C_m$.
Then, the confusion score of the class $C_m$ can be reformulated as:
\begin{equation}\label{error_score}
\begin{array}{ll}
S_{dist}(C_m) &= \frac{1}{N_m}\sum\limits_{(x, y) \in C_m}\sum\limits_{1 \leq j \leq M} \frac{||g^m_x - u^m||^2}{||g^m_x - u^j||^2}\\
&=1 + \frac{1}{N_m}\sum\limits_{(x, y) \in C_m}\sum\limits_{1\leq j \leq M, j\ne m} \frac{||g^m_x - u^m||^2}{||g^m_x - u^j||^2},
\end{array}
\end{equation}
where $||\cdot||^2$ denotes the squared euclidean distance. 

\noindent\textbf{Entropy-based Criterion.} This criterion is formulated as:
\begin{equation}\label{error_score_entropy}
\begin{array}{ll}
S_{entropy}(C_m) &= \frac{1}{N_m}\sum\limits_{(x, y) \in C_m} p^m_xlog\frac{1}{p^m_x},\\
\end{array}
\end{equation}
where $S_{entropy}(C_m)$ denotes the confusion score of $C_m$.

We can observe that the above two confusion criteria measure the confusion score in different spaces. The confusion score obtained by the distance-based criterion is measured in feature space, which rises as the features in a certain class move away from the center of that class and approach other class centers. The confusion score obtained by the entropy-based criterion is measured in logits space, which rises when the logits of samples of a certain class move away from the one-hot vector.

Based on the obtained scores for each class, we can get an ordered dataset:
\begin{equation}\label{order_dataset}
\begin{array}{c}
D_{ord}=C_{{ord}_1} \cup C_{{ord}_2} \cup \dots \cup C_{{ord}_M},
\end{array}
\end{equation}
where ${ord}_m$ is the index of the class with the $m$-th largest confusion score. The sorting process is detailed in \cref{fig:order}.

\begin{figure}[t]
	\centering
	\includegraphics[width=1\columnwidth]{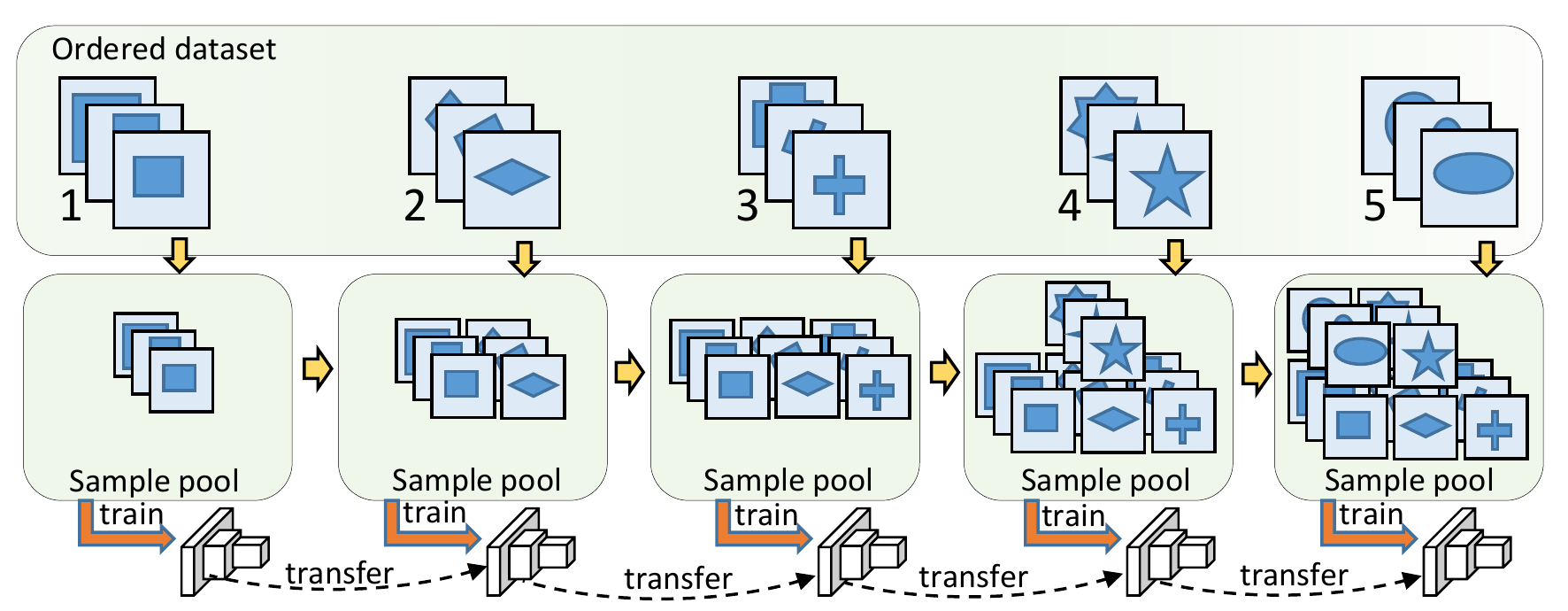}
	\caption{Illustration of our class-based expansion learning method. We first learn the network by a training sample pool which only contains a small portion of the classes from an ordered dataset. Based on the previously learned model, we learn the network by the class-based expanded training sample pool with some classes newly added from the ordered dataset. We repeat this process after all the classes of the ordered dataset are added to the training sample pool.}
	\label{fig:pde}
\end{figure}

\subsection{Progressive Expansion Learning}\label{expansion}
We now describe our proposed progressive expansion learning pattern for CNN models. 
With the ordered dataset $D_{ord}$, we split the optimization of \cref{func} into $K$ stages (For convenience, we set $M$ to be divisible by $K$). We start with an empty training sample pool ($D_{ord}^0=\emptyset$). At the first stage, the first $\frac{M}{K}$ classes from the ordered dataset $D_{ord}$ are added to $D_{ord}^0$, then the training sample pool is expanded to $D_{ord}^1$:
\begin{equation}\label{subdataset_1}
\begin{array}{ll}
D_{ord}^1=C_{{ord}_1} \cup C_{{ord}_2} \cup \dots \cup C_{{ord}_{\frac{M}{K}}}.
\end{array}
\end{equation}
The target optimization function of $D_{ord}^1$ is:
\begin{equation}\label{sunfunc_1}
\begin{array}{ll}
\theta_1^* = \mathop{\arg\min}\limits_{\theta} \sum\limits_{(x, y) \in D_{ord}^1}l(f(x;\theta), y),
\end{array}
\end{equation}
where $\theta$ is randomly initialized and $\theta_1^*$ represents the optimal model parameters learned from $D_{ord}^1$. At the $k$-th stage ($1 < k \leq K$), the training sample pool $D_{ord}^{k-1}$ is expanded to $D_{ord}^k$:
\begin{equation}\label{subdataset}
\begin{array}{ll}
D_{ord}^k=C_{{ord}_1} \cup C_{{ord}_2} \cup \dots \cup C_{{ord}_{\frac{kM}{K}}},
\end{array}
\end{equation}
where the last $\frac{M}{K}$ classes of $D_{ord}^k$ are newly added. In order to find the optimal model parameters $\theta_k^*$ for $D_{ord}^k$, we have:
\begin{equation}\label{sunfunc}
\begin{array}{ll}
\theta_k^* = \mathop{\arg\min}\limits_{\theta} \sum\limits_{(x, y) \in D_{ord}^k}l(f(x;\theta), y),
\end{array}
\end{equation}
where the $\theta$ is initialized by the optimal model parameters $\theta_{k-1}^*$ learned from $D_{ord}^{k-1}$.


In the simplest form of class-based expansion learning, the classes of the dataset are progressively added to the training sample pool. By analogy, using such a progressive way, we will eventually solve the problem in \cref{func} after the samples of all the classes participate in the training process. 

\subsection{Complexity Analysis}\label{time}
In this section, we consider the time complexity of class-based expansion learning (CEL). Let $T_{normal}$ be the time cost for a normal training process. For CEL, if we use the same number of epochs for each stage, the time cost will be $\frac{kT_{normal}}{K}$ at stage $k$ (the ratio of the dataset size of stage $k$ to the size of the entire dataset is $\frac{k}{K}$). Then the class-based expansion learning time cost $T_{CEL}$ is:
\begin{equation}\label{epochs_time}
\begin{array}{ll}
T_{CEL} &= \sum_{k=1}^{K} \frac{kT_{normal}}{K} = \frac{(K+1)}{2}T_{normal},
\end{array}
\end{equation}
which is a linear time algorithm. 

In the experiment, we observe that reducing the number of epochs in the early stages by a factor of $\lambda$ does not sacrifice accuracy. We then train the network with the full amount of epochs only at the final stage and reducing the epoch number in other stages. In this way, the time cost $T_{CEL2}$ is:
\begin{equation}\label{epochs_number_2}
\begin{array}{ll}
T_{CEL2} = (\sum_{k=1}^{K-1} \frac{k}{\lambda K} + 1) T_{normal}= (\frac{(K-1)}{2\lambda} + 1)T_{normal}.
\end{array}
\end{equation}
We can reduce the time cost of CEL by controlling the value of $\lambda$. With a large $\lambda$, only a little consumption time is required at the early stage, making the training time of our strategy comparable to the training time of normal training.

\section{Experiments}
\subsection{Experimental Settings}

\begin{table*}[t]
	\centering
	\caption{Test errors (\%) of Normal Training and CEL for CIFAR10.}	\resizebox{1\textwidth}{!}{
		\begin{tabular}{l|l|c|c|cccccccccc}
			\hline
			\multirow{2}*{Method} & \multirow{2}*{Network} & \multirow{2}*{Runs} & \multirow{2}*{Test error (\%)} &\multicolumn{10}{c}{Test error of each class (\%)}\\
			\cline{5-14}
			&&& & Airplane & Automobile & Bird & Cat & Deer & Dog & Frog & Horse & Ship & Truck\\
			\hline
			\multirow{2}*{Normal Training}& ResNet-32 & 5 & 7.08 & 6.00 & 3.50 & 9.60 & 14.40 & 5.70 & 11.90 & \textbf{4.70} & \textbf{4.90} & 5.00 & 4.80\\
			&ResNet-110 & 5 & 6.24 & \textbf{4.40} & \textbf{2.40} & 9.70 & 12.40 & 4.70 & 12.00 & 3.70 & \textbf{4.50} & \textbf{4.10} & \textbf{4.50}\\				
			\hline
			\multirow{2}*{CEL}&ResNet-32 & 5 & \textbf{6.16} & \textbf{4.50} & \textbf{3.00} & \textbf{7.80} & \textbf{12.50} & \textbf{4.30} & \textbf{11.60} & 4.90 & 5.10 & \textbf{3.40} & \textbf{4.50}\\
			&ResNet-110 & 5 & \textbf{5.71} & 4.90 & 2.60 & \textbf{7.40} & \textbf{11.80} & \textbf{4.30} & \textbf{9.60} & \textbf{2.90} & 4.60 & 4.30 & 4.70\\
			\hline
		\end{tabular}
		
	}
	\label{tab:cifar10}%
	\vspace{-10pt}
\end{table*}%

\begin{table}[!h]
	\centering
	\caption{Test errors (\%) of different methods for CIFAR10 and CIFAR100 based on ResNet-32.}
	\huge
	\vspace{-8pt}
	\resizebox{1\columnwidth}{!}{
		\begin{tabular}{l|c|c|c|c|c|c}
			\hline
			Dataset& Normal Training & CBS~\cite{sinha2020curriculum} & DIHCL~\cite{zhou2020curriculum} & Curriculum~\cite{wu2021when} & CEL& CEL-2\\
			\hline
			CIFAR10& 7.08 & 7.99 & 6.93 & 6.88 & \textbf{6.16} & 6.84 \\
			\hline
			CIFAR100& 30.40 & 31.49 & 31.47 & 30.02 & 29.82 & \textbf{29.41}\\
			\hline
		\end{tabular}
	}
	\label{tab:sota}%
	\vspace{-18pt}
\end{table}%

\begin{table}[t]
	\centering
	\tiny
	\caption{Test errors (\%) of normal training and CEL for ImageNet100.}
	\resizebox{1\columnwidth}{!}{
		\begin{tabular}{l|l|c|c}
			\hline
			Dataset&Network& Normal Training & CEL\\
			\hline
			ImageNet100&ResNet-18& 29.83 & \textbf{26.86}\\
			\hline
		\end{tabular}
	}
	\label{tab:cifar100}%
	\vspace{-10pt}
\end{table}%
\begin{table}[!t]
	\centering
	\caption{The class order of the ordered dataset for CIFAR10 based on our class confusion criterion.}
	\scriptsize	\resizebox{1\columnwidth}{!}{
		\begin{tabular}{l|cccccc}
			\hline
			Ranking& 1 & 2 & 3 & 4 & 5\\
			\hline
			class name & Cat & Bird & Dog & Airplane & Deer\\
			\hline
			Ranking& 6 & 7 & 8 & 9 & 10\\
			\hline
			class name & Frog & Horse & Truck & Ship & Automobile\\
			\hline
		\end{tabular}
		
	}
	\label{tab:class_order}%
	\vspace{-10pt}
\end{table}%

\paragraph{Dataset} We conduct our experiments on three datasets, namely CIFAR10, CIFAR100, and ImageNet100. The CIFAR10~\cite{krizhevsky2009learning} dataset is a labeled subset of the 80 million tiny images dataset~\cite{4531741}, which consists of 60,000 RGB images of resolution 32$\times$32 in 10 classes, with 5,000 images per class for training and 1,000 per class for testing. The CIFAR100~\cite{krizhevsky2009learning} is similar to the CIFAR10, except that it has 100 classes containing 600 images each.	There are 500 training images and 100 testing images for each class of CIFAR100. The ImageNet100 is a subset of ImageNet~\cite{krizhevsky2012imagenet} for ImageNet Large Scale Visual Recognition Challenge 2012. It contains 129,395 training images and 5,000 validation images in the first 100 classes of ImageNet.

\paragraph{Data preprocessing} On CIFAR10 and CIFAR100, we just follow the simple data augmentation in ResNet~\cite{he2016deep} for training, including random cropping for 4 pixels padded image, per-pixel mean subtraction and horizontal flip. On ImageNet100, the augmentation strategies we use are the 224$\times$224 random cropping and the horizontal flip.

\paragraph{Implementation details} We conduct our class-based expansion learning scheme on CIFAR10, CIFAR100, and ImageNet100 by using the state-of-the-art CNN models, including ResNet-18, ResNet-32 and ResNet-110. On CIFAR10, as described in \cref{criterion}, we use a pre-trained ResNet-20 (trained by 60 epochs) on ImageNet to determine the order of classes. Then, as described in \cref{expansion}, we divide the learning of the ordered dataset into $5$ stages. At the first four stages, we use $60$ epochs to train the network and we train the network with $300$ epochs at the last stage, i.e., $\lambda = K = 5$. At each stage, we train the network using SGD with a mini-batch size of 128, a weight decay of $0.0001$ and a momentum of $0.9$. The initial learning rate is set to $0.1$ and is divided by $10$ after $\frac{1}{2}$ and $\frac{3}{4}$ of all epochs. On CIFAR100, we also use the pre-trained ResNet-50 to determine the order of classes. Afterwards, we divide the learning of the ordered dataset into $10$ stages. At the first nine stages, we utilize $60$ epochs to train the network and we train the network with $200$ epochs at the final stage. The other parameters of the experiments are the same as those used on CIFAR10. On ImageNet100, we use a pre-trained ResNet-18 to determine the order of classes (with 30 epochs). We divide the learning of the dataset into $10$ stages by the original order. At the first nine stages, we use 60 epochs to train the network and we train the network with $60$ epochs at the final stage. The initial learning rate is set to 2 and is divided by 5 after 20, 30, 40 and 50 epochs. The rest of the settings are the same as those on CIFAR10. We implement our scheme with the theano~\cite{bergstra2010theano} and use an NVIDIA TITAN 1080 Ti GPU to train the network.

\subsection{Comparisons with the State of the Art}
We compare our class-based expansion learning (CEL) with other state-of-the-art methods of Normal Training, CBS~\cite{sinha2020curriculum}, DIHCL~\cite{zhou2020curriculum}, and Curriculum~\cite{wu2021when}. The normal training method represents a standard training method, the CEL adopts the distance-based confusion criterion, and the CEL-2 adopts the entropy-based class confusion criterion. The results are summarized in \cref{tab:sota}. As shown in \cref{tab:sota}, CEL and CEL-2 outperform other state-of-the-art methods, which demonstrates their effectiveness.

We also employ ResNet-18 to conduct experiments on ImageNet100. The results are summarized in \cref{tab:cifar100}. Similar conclusions to those on CIFAR10 datasets can be made. These results demonstrate the generalization of our approach.

\begin{figure}[t]
	\centering
	\subfigure[Training loss]{
		\begin{minipage}[ht]{0.30\columnwidth}
			\includegraphics[width = 1\columnwidth]{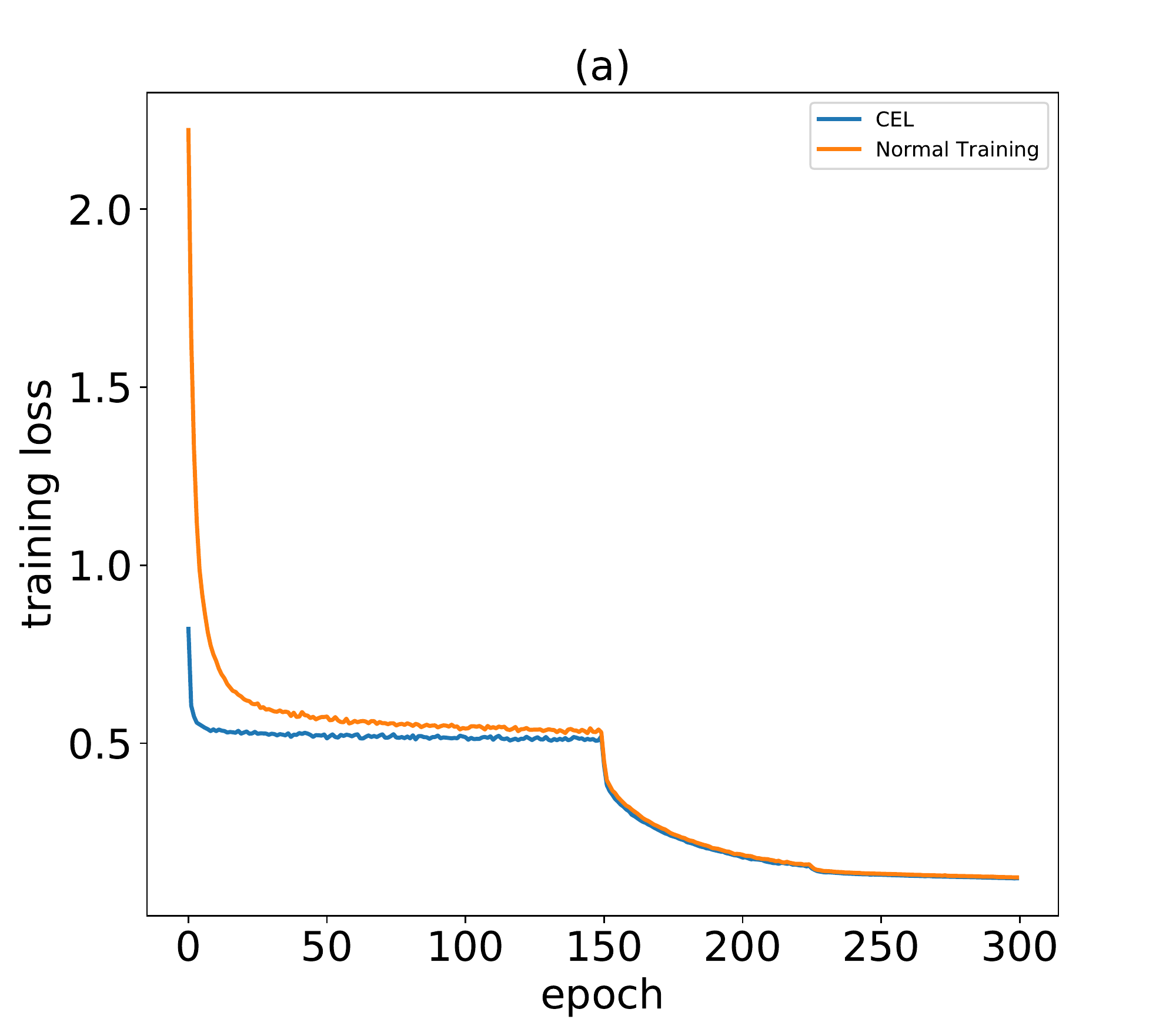}	
	\end{minipage}}
	\subfigure[Validation loss]{
		\begin{minipage}[ht]{0.30\columnwidth}
			\includegraphics[width = 1\columnwidth]{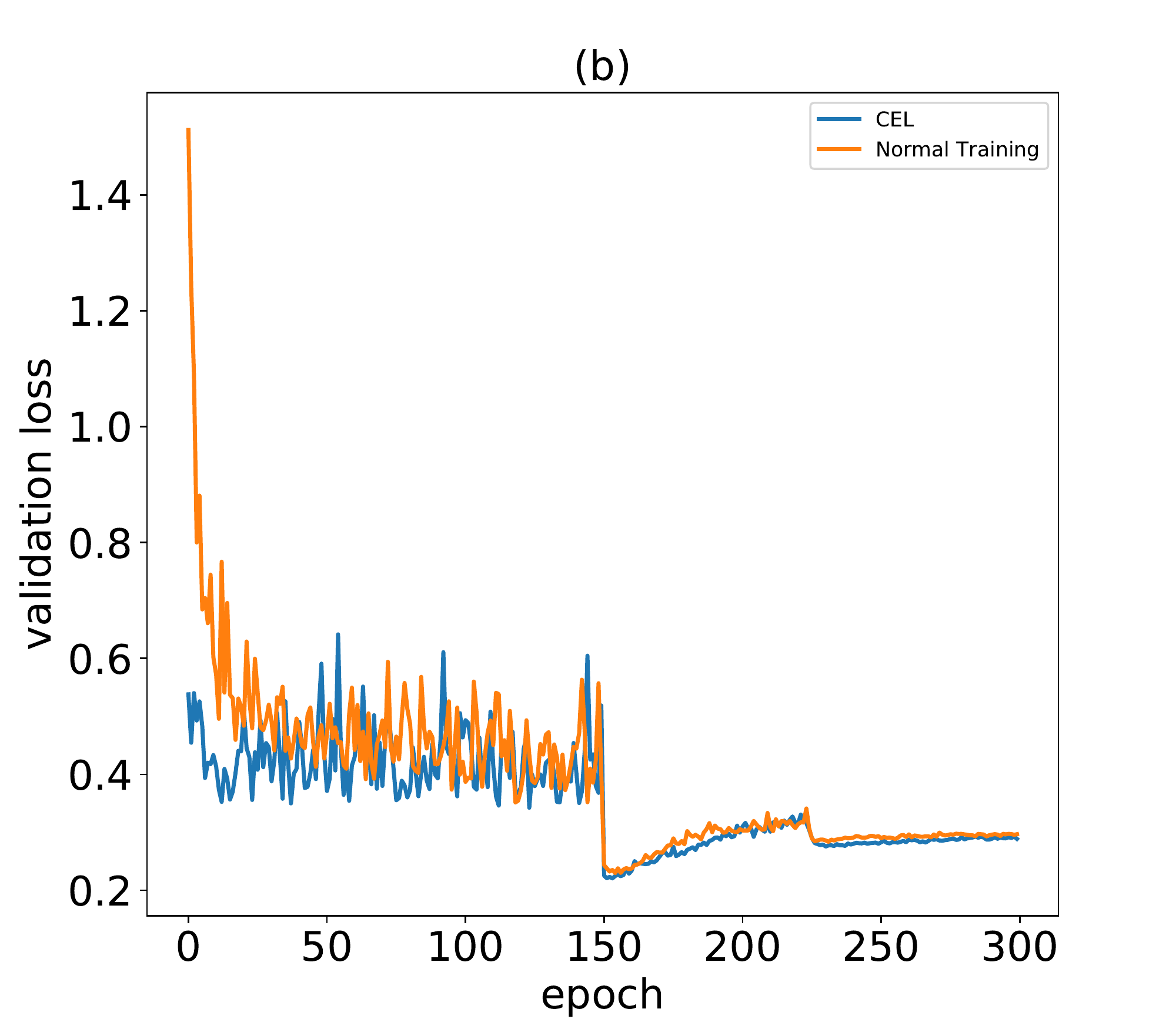}
		\end{minipage}	
	}
	\subfigure[Accuracy]{
		\begin{minipage}[ht]{0.30\columnwidth}
			\includegraphics[width = 1\columnwidth]{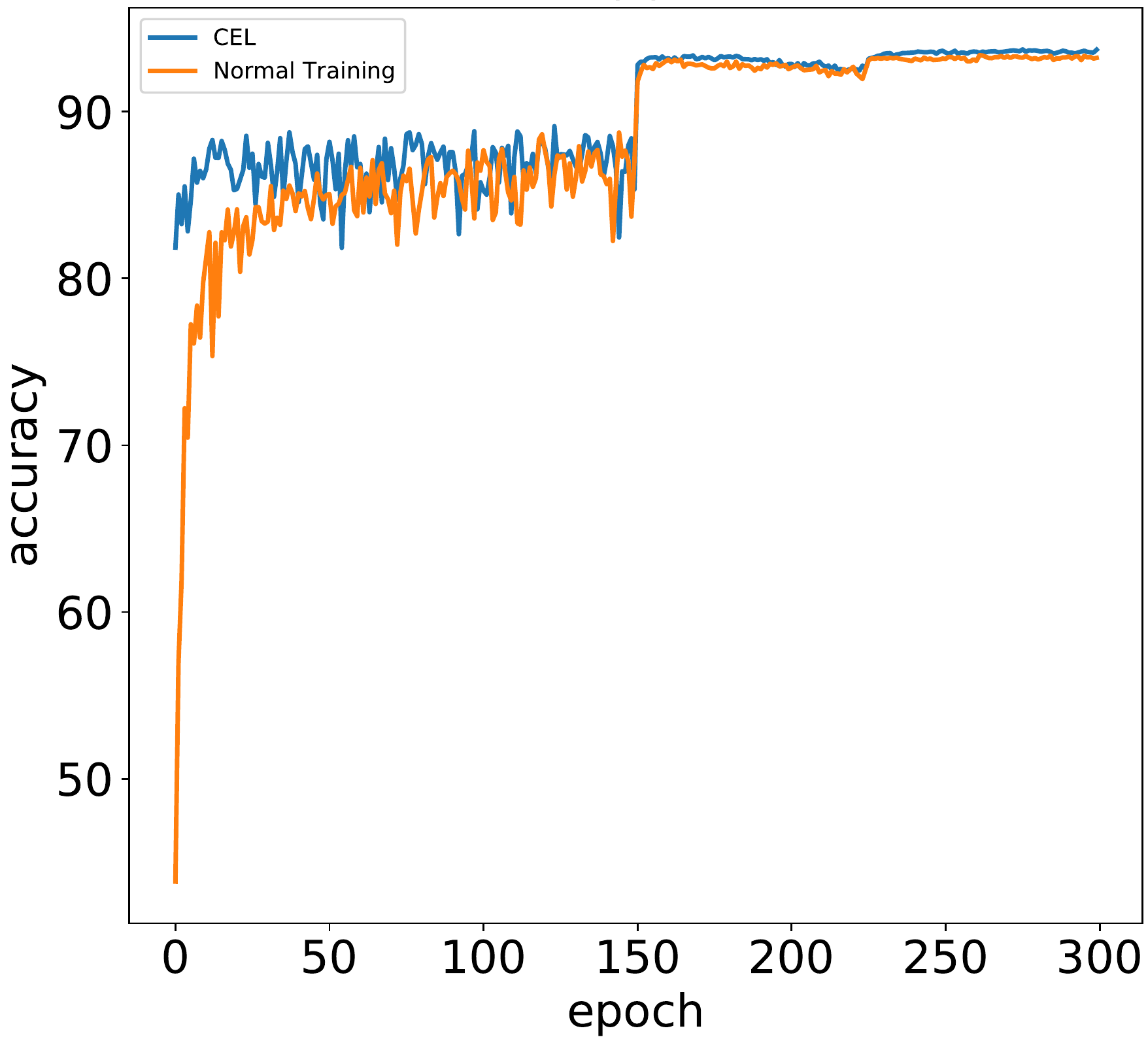}	
		\end{minipage}
	}

	\caption{The final stage of CEL method and normal training for CIFAR10 based on ResNet-32.}
	\label{fig:convergence}
	\vspace{-10pt}
\end{figure}

\subsection{Ablation Experiments}

\paragraph{Analysis of the class order} We presented the test error results for each class of CIFAR10 in \cref{tab:cifar10} and the class order in \cref{tab:class_order}. As shown in \cref{tab:class_order}, we can observe the cat class, the bird class, as well as the dog class, are both confusing classes defined by the distance-based confusion criterion, and the error rates of these classes are the largest ones in \cref{tab:cifar10}. \cref{tab:cifar10} gives the results on CIFAR10, which shows that our method outperforms the normal training method on ResNet-32 and ResNet-110. In addition, we observe that the improvement in the performance of the model was mainly due to the preferentially selected classes (i.e. cat, bird, deer, dog, and airplane).

\paragraph{Analysis of the individual components} We carry out an experiment on CIFAR10 to analyze the individual components in the CEL method. In this experiment, without the sorted class order obtained by $g(\cdot)$, we perform class expansion learning in a random class order, which is denoted by ``w/o $g(\cdot)$''. The results in \cref{tab:order} indicate that ``w/o $g(\cdot)$'' performs better than normal training due to the class-based expansion learning process. In addition, ``w/ $g(\cdot)$'' also can improve the performance of ``w/o $g(\cdot)$'', showing the effectiveness of the sorted class order obtained by $g(\cdot)$.

\begin{table}[t]
	\centering
	\caption{Test errors (\%) of different methods for CIFAR10 and CIFAR100 based on ResNet-32.}
	\tiny
	\resizebox{1\columnwidth}{!}{
		\begin{tabular}{l|c|c|c}
			\hline
			Dataset & Normal Training&w/ $g(\cdot)$& w/o $g(\cdot)$\\
			\hline
			CIFAR10& 7.08 & \textbf{6.16} & 6.32\\
			\hline
			CIFAR100&30.40 & \textbf{29.82} & 30.16\\
			\hline
		\end{tabular}
	}
	\vspace{-10pt}
	\label{tab:order}%
\end{table}%

\begin{table}[!t]
	\centering
	\caption{Test errors (\%) of different methods at the same number of epoch times for CIFAR100 and ImageNet100. One epoch time is the time of traversing the entire training dataset once.}
	\huge
	\resizebox{1\columnwidth}{!}{
		\begin{tabular}{l|c|c|c|c|c}
			\hline
			\multirow{2}*{Dataset}&\multirow{2}*{Network}& \multicolumn{2}{|c|}{Normal Training} & \multicolumn{2}{|c}{CEL}\\
			\cline{3-6}
			& &Epoch time & Test error & Epoch time & Test error\\
			\hline
			CIFAR10&ResNet-32& 420 & 6.50 & 420 & \textbf{6.16}\\
			\hline
			ImageNet100&ResNet-18& 330 & 27.52 & 330 & \textbf{26.86}\\
			\hline
		\end{tabular}
	}
	\label{tab:epoch_time}%
	\vspace{-18pt}
\end{table}%

\paragraph{Convergence performance of final stage} The convergence performance of the final stage is shown in \cref{fig:convergence}. From \cref{fig:convergence}, we observe that our method converges faster than the normal training at the beginning, and performs better in most cases. These observations mean that learning local classes in advance can effectively accelerate network convergence.

\paragraph{Impact of long time training} To evaluate the impact of the long time training, we conduct the experiments on CIFAR10 and ImageNet100 where we make the time cost of the normal training the same as the one of CEL. In these experiments, we increase the number of epochs in normal training to match the one used in the CEL method. \cref{tab:epoch_time} gives the results, which show that our method outperforms the normal training method at the same number of epochs.

\section{Conclusion}
In this letter, we have presented a novel class-based expansion learning scheme for CNN, which learns the whole dataset by progressively training the CNN model in a bottom-up class growing manner. By using this scheme, the classification boundaries of the preferentially selected classes are frequently stimulated, resulting in a fine-grained form. Based on the characteristics of the scheme, we have also proposed a class confusion criterion that prioritizes the classes that are easily confused. Extensive experimental results demonstrate the effectiveness of our work.

\bibliographystyle{IEEEtran}
\bibliography{IEEEtran}
\end{document}